\documentclass[letterpaper, 10 pt, conference]{ieeeconf}

\IEEEoverridecommandlockouts                              

\usepackage{amsmath}
\usepackage{amssymb}
\usepackage[ruled]{algorithm}
\usepackage{algpseudocode}
\usepackage[table]{xcolor} 
\usepackage{multirow}
\usepackage{stix}
\usepackage{array}
\newcolumntype{P}[1]{>{\centering\arraybackslash}p{#1}}
\usepackage{tikz}
\usepackage[hidelinks]{hyperref}
\usepackage{cite}

\title{\LARGE \bf
An Integrated Dynamic Method for Allocating Roles and Planning Tasks for Mixed Human-Robot Teams
}

\author{Fabio Fusaro$^{1,2*}$, Edoardo Lamon$^{1*}$, Elena De Momi$^{2}$, and Arash Ajoudani$^1$
\thanks{$^*$ Contributed equally to this work.}
\thanks{1- HRI$^2$ Lab, Istituto Italiano di Tecnologia, Via  Morego  30, 16163, Genova, Italy. {2- Department of Electronics, Information and Bioengineering, Politecnico di Milano, Milano, Italy.} {\tt\small fabio.fusaro@iit.it}}
\thanks{This work was supported by the European Research Council's (ERC) starting grant Ergo-Lean (GA 850932).}
}

\begin{document}

\maketitle
\thispagestyle{empty}
\pagestyle{empty}

\begin{abstract}
This paper proposes a novel integrated dynamic method based on Behavior Trees for planning and allocating tasks in mixed human robot teams, suitable for manufacturing environments. 
The Behavior Tree formulation allows encoding a single job as a compound of different tasks with temporal and logic constraints. In this way, instead of the well-studied offline centralized optimization problem, the role allocation problem is solved with multiple simplified online optimization sub-problem, without complex and cross-schedule task dependencies. These sub-problems are defined as Mixed-Integer Linear Programs, that, according to the worker-actions related costs and the workers' availability, allocate the yet-to-execute tasks among the available workers.
To characterize the behavior of the developed method, we opted to perform different simulation experiments in which the results of the action-worker allocation and computational complexity are evaluated. The obtained results, due to the nature of the algorithm and to the possibility of simulating the agents' behavior, should describe well also how the algorithm performs in real experiments.
\end{abstract}

\section{INTRODUCTION}
The increasing demand for flexible and highly reconfigurable production lines of small-medium size enterprises needs industrial manipulators to be able to quickly adapt to diversified manufacturing requirements.
In this context, torque-controlled collaborative robots (cobots), are not only able to deal with complex tasks~\cite{ajoudani2018progress} and to execute safe plans in human-populated and partially unstructured environments~\cite{haddadin2008collision}, but also to offload workers from repetitive and hard tasks~\cite{kim2018anticipatory}.

Moreover, cobots are expected to be able to perform a wide variety of tasks, both autonomously and in collaboration with human co-workers, that can supervise and complement robot performance with superior expertise and task understanding, assembling proper mixed human-robot teams. This scenario reveals two fundamental problems: how to systematically assign a role to each member of the team to achieve a shared goal, and how to dynamically adapt the robot plan to dynamical changes of role between the team agents. 

In the literature, a common approach to model the problem of allocating roles in a team of agents is through combinatory approaches. A formal analysis, comprehensive of computational models, is presented first by \textit{Gerkey and Matari\'c} \cite{gerkey2004formal} and then updated by \textit{Korsah et al.} \cite{korsah2013comprehensive}. Examples of such methods, applied to human-robot mixed teams are represented by \cite{gombolay2015coordination, ferreira2011scheduling}. In order to enable the algorithm with dynamic behaviors, that can adapt the offline plan with the online contingencies, researchers started to add also dynamic re-scheduler \cite{pupa2021human}.
Another extensively used method in human-robot teams is represented by AND/OR graphs, for the ability to decompose assembly tasks~\cite{johannsmeier2017hierarchical, darvish2021hierarchical}.
An online scheduler based on time Petri nets is developed in~\cite{casalino2019optimal}, where the robot adapts its schedule based on human activities. In this direction, researchers started to study also which features should be considered in such problems, adapting the robot plan to human co-worker preferences, capabilities and ergonomics~\cite{gombolay2017computational, michalos2018method, lamon2019capability, el2019task}. 

\begin{figure}[t]
    \centering
    \includegraphics[trim=0.0cm 0.0cm 0cm 0.cm,clip,width=\linewidth]{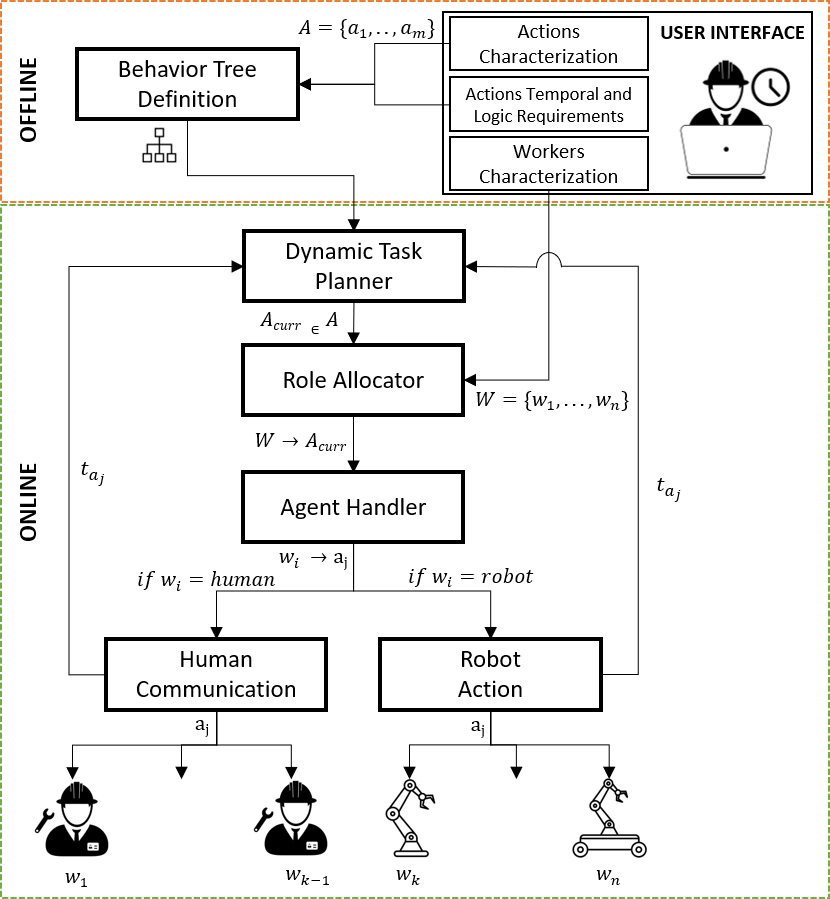}
    \caption{Scheme of the allocation and planning method. First from the user interface, actions and workers information are defined offline. Next, the BT is constructed using the received characterizations. Then, in the online phase, the BT starts its execution and dynamically allocate the actions to the workers, exploiting the \textit{Role Allocator}. Finally, through the \textit{Agent Handler}, the BT either communicates the action to the human worker or makes the robot executing the task.}
    \label{fig:bt_allocation_scheme}
    \vspace{-7mm}
\end{figure}

The main limitation of these approaches consists in the fact that the task allocation and planning problems are solved in two different phases, often using two completely different and separated methods. The proposed method, instead, tries to exploits the strength of a centralized reactive and modular task planning method, in charge of scheduling the tasks executions, with the advantages of a cost-based role allocator, that, before the proper execution phase, solves dynamically the problem of allocating only a subset of actions to the agents. The role of the task planner is to dynamically schedule different tasks, ensuring that the task temporal and logic constraints, in terms of sequence or parallel of tasks, are satisfied. In this way, the planner simplifies the centralized problem of allocating all the tasks to all the agents, to the sub-optimal, but computationally less expensive, problem of allocating only a subset of tasks to the agents, without any constraint generated by the plan. 
In particular, the centralized task planner is defined by means of Behavior Trees (BTs), which triggers the action execution of all the agents of the team. Differently from the usual approach, where the behavior of each robotic agent is ruled by its own planner, here the BT models the task instead of the agent. The main advantage consists in the fact that, in this way, it can delegate each task to different agents, according to the allocation results generated by the \textit{Role Allocation} node. In particular, thanks to the \textit{Agent Handler} node, the tasks assigned to the robotic agents will be directly executed through the \textit{Action} nodes, whereas the other ones will inform the human workers by means of the \textit{Human Communication} node. The architecture of the method is depicted in \autoref{fig:bt_allocation_scheme}. 
On the other hand, the role allocation problem allows to dynamically assign a suitable agent for each task. The suitability is evaluated through the action-worker related costs, which are used to describe how good an agent is good in performing a task. These costs should model both general kino-dynamic agent features, tasks duration, availability, and also more specific ones, such as expertise and ergonomics. In particular, by taking into account tasks' duration and agent availability, it is also possible to dynamically schedule the tasks, to coordinate agents' effort.
The method has been tested with simulation experiments in which both the proposed BT structure and the role allocation problem are evaluated. First, the computation complexity of the whole framework is computed increasing progressively the number of actions and workers. Then, the allocation algorithm is compared with different values of the agent-related availability cost.

To summarize, the contribution of this manuscript consists of a generic method to manage complex decomposable jobs with schedule constraints, able to optimally assign an agent and plan every single task of the job in a heterogeneous human-robot team. To enable the task planner to dynamically decide which agent is the most suitable to allocate the task, we created four BT nodes to solve the role allocation for the sub-tasks and the consequent agents' handler and executor. 

\section{MODIFIED BTs with ROLE ALLOCATION}
\subsection{Preliminaries on Behavior Trees}
A BT is a directed rooted tree, consisting of internal nodes for control flow and leaf nodes for action execution or condition evaluation. It is composed by  \textit{parent} and \textit{child} nodes that are the adjacent pair. The main node, root, is the only one without parents and starts the execution of its children propagating a signal, called \textit{tick}, through the tree. Once the children are ticked, they return immediately to the parent a status: if the child is executing returns \textit{RUNNING}, if the node completes the execution successfully, returns \textit{SUCCESS}, otherwise, it fails and returns \textit{FAILURE}. There are two main types of nodes: control and execution. The control ones are divided into four standard categories (Sequence, Fallback, Parallel, Decorator), while the execution ones in two (Action, Condition). The standard types of nodes, with their symbol and the return status depending on each case, are summarized in \autoref{table:BT_nodes}.
 
Standard BTs were designed to control an agent behavior, by reactively plan tasks to execute~\cite{colledanchise2017behavior}. Thanks to their design, they allow generating different behaviors that satisfy conditions evaluated online. Moreover, it also envisions the execution of tasks both in parallel and in sequence, it is possible to adapt it in human-robot cooperative tasks, such as assembly, to achieve industrial jobs.

\begin{table}[!t]
\caption{BT node types and return status.}
\label{table:BT_nodes}
\vspace{-8mm}
\title{BT node types and return status.}
\begin{center}
\resizebox{\columnwidth}{!}{\begin{tabular}{|P{.2\columnwidth}|P{.2\columnwidth}|P{.2\columnwidth}|P{.2\columnwidth}|P{.2\columnwidth}|}
\hline
\multirow{2}{*}{\textbf{Type of Node}} & \multirow{2}{*}{\textbf{Symbol}} & \multirow{2}{*}{\textbf{Success}} & \multirow{2}{*}{\textbf{Failure}} & \multirow{2}{*}{\textbf{Running}} \\[3mm]
\hline
\end{tabular}}

\vspace*{0.1 cm}

\resizebox{\columnwidth}{!}{\begin{tabular}{|P{.2\columnwidth}|P{.2\columnwidth}|P{.2\columnwidth}|P{.2\columnwidth}|P{.2\columnwidth}|}
\hline
\multirow{3}{*}{Sequence} & \multirow{3}{*}{$\rightarrow$} &\multirow{2}{*}{All children} \multirow{2}{*}{succeed}  & \multirow{3}{*}{One child fails} & One child returns Running \\
\hline
\multirow{3}{*}{Fallback} & \multirow{3}{*}{?} & \multirow{2}{*}{One} \multirow{2}{*}{child succeeds} & \multirow{2}{*}{All children} \multirow{2}{*}{fail} & One child returns Running \\
\hline
\multirow{2}{*}{Decorator} & \multirow{2}{*}{$\diamondsuit$} & \multirow{2}{*}{Custom} & \multirow{2}{*}{Custom} & \multirow{2}{*}{Custom} \\[3mm]
\hline
\multirow{2}{*}{Parallel} & \multirow{2}{*}{$\rightrightarrows$} & $\geq M$ children succeed & $> N - M$ children fail & \multirow{2}{*}{else}\\
\hline
\end{tabular}}

\vspace*{0.1 cm}

\resizebox{\columnwidth}{!}{\begin{tabular}{|P{.2\columnwidth}|P{.2\columnwidth}|P{.2\columnwidth}|P{.2\columnwidth}|P{.2\columnwidth}|}
\hline
\multirow{2}{*}{Condition} & \multirow{2}{*}{\tikz \draw (0,0) ellipse (4pt and 2pt);} & \multirow{2}{*}{True} & \multirow{2}{*}{False} & \multirow{2}{*}{Never}\\[3mm]
\hline
\multirow{2}{*}{Action} & \multirow{2}{*}{$\hrectangle$} & Upon completion  & Impossible to complete & During execution\\
\hline
\end{tabular}}
\end{center}
\vspace{-7mm}
\end{table}

\subsection{Nodes for Role Allocation}

To embed the role allocation problem in a BT, the standard usage of the method should have modified. First, it is important to specify that, in our method, the BT controls the job behavior, instead of the agent behavior, where a job is represented by the set of its tasks with their temporal constraints. In this way, we exploit the BT to model all the possible execution of a job, and only when one or multiple jobs should be executed, the role allocation problem is solved and finally, the different agents are informed of the results (see \autoref{fig:bt_allocation_scheme}). 
To do so, we defined four custom nodes and a particular subtree is designed by the means of the standard types of nodes modifying their functionality. 

Each developed node has to communicate with the others to share information and data. When the nodes are not directly connected or the return values are not sufficient to achieve desired behaviors, the BT exploits input/output ports. Each port is defined by a unique name and can be used as static to read elements that are input from the external in the creation phase of the BT or as dynamic to read and/or write data. The output ports write the elements into a shared \textit{blackboard} associating to each variable a name. The input port can access the data knowing the name of the variable.

\begin{figure}
    \centering
    \includegraphics[trim=0.0cm 0.3cm 0cm 0.8cm,clip,width=0.7\linewidth]{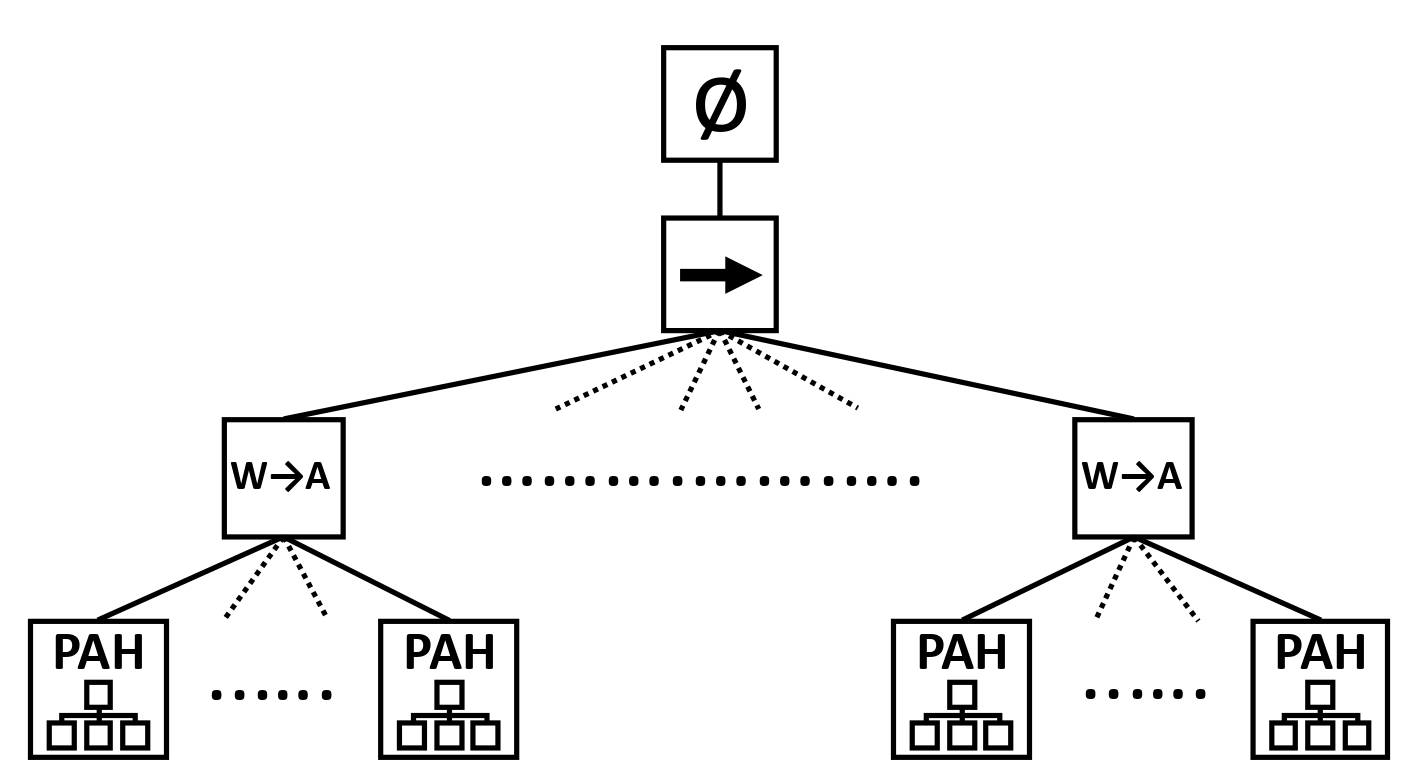}
    \caption{Planning and Allocation Behavior Tree. The structure of the BT is composed by a sequence of \textit{Role Allocator} nodes ($W \rightarrow A$). Each allocator node has the \textit{Planning and Allocation Handler} subtree (PAH), shown in \autoref{fig:subtree_alloc}, as children.}
    \label{fig:tree_alloc}
    \vspace{-5mm}
\end{figure}

\begin{figure}
    \centering
    \includegraphics[trim=0.0cm 0.0cm 0cm 0.40cm,clip,width=0.7\linewidth]{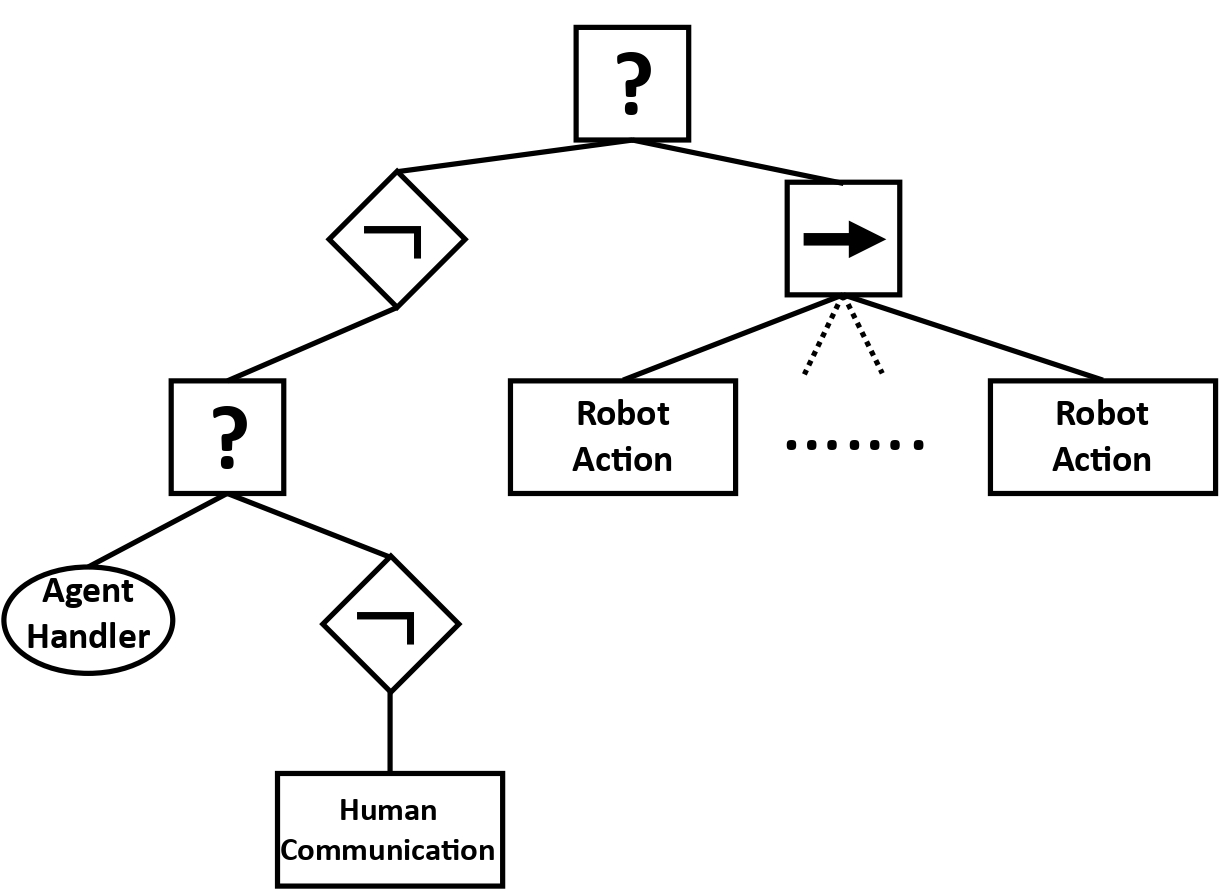}
    \caption{Planning and Allocation Handler subtree. The developed structure, using two fallbacks (?) and two inverters ($\neg$) allows to manage the return status of the \textit{Agent Handler} node to either communicate the allocation to the human or make the robot executing the sequence of primitive actions.}
    \label{fig:subtree_alloc}
    \vspace{-5mm}
\end{figure}

The general structure of the BT developed to plan and allocate tasks is shown in \autoref{fig:tree_alloc}. 
The tree shows the combination of tasks in series and parallel, that ends with a sequence. Before the execution layer, i.e. the action nodes, we designed a fixed subtree where the \textit{Role Allocator} nodes, represented by the symbol $W \rightarrow A$, is in charge of solving the allocation of such actions to the agents and the subtree \textit{Planning and Allocation Handler}, children of the Role Allocator displayed in \autoref{fig:subtree_alloc}, is in charge of delegating the task to each agent. Each allocator node has a number of children equal to the number of tasks that can be executed in parallel, that can be one or multiple.
The custom nodes are explained in details in the following subsections. 

\subsubsection{Role Allocator Node}
This node is the parent of the \textit{Planning and Allocation Handler} subtree. It is defined as a control node and shares few similarities with the parallel node.
The children of this node are the \textit{Planning and Allocation Handler} subtrees, one for each task that can be executed in parallel. The node reads from the static input port the info of the children's actions that still have to be executed, and the agents' info. In this way, the node can calculate all the agents-task related costs, generate the role allocation problem as explained in \autoref{sec:role_allocation}, and outputs the result. 
Then, each allocated action ticks the related child.
The node is then executed again until all the tasks are completed.
The pseudocode of the \textit{Role Allocator} node is synthesized in Algorithm~\ref{algo:RoleAllocator}.

\alglanguage{pseudocode}
\begin{algorithm}[!ht]
\small
\caption{Tick() function of the "RoleAllocator" node.}
\label{algo:RoleAllocator}
\begin{algorithmic}[1]
\Procedure{$\textproc{RoleAllocator::tick}$}{$ $}
\If {$action\_to\_be\_allocated \neq 0$}
    \State $[W,\;A] = Allocate()$
    \State $setOutput([W,\;A])$
\EndIf
\For{$[w,\;a]$ \textbf{in} $[W,\;A]$}
    \State $child\_idx \gets a.ID$
\EndFor
\For{$idx$ \textbf{in} $child\_idx$}
    \If {$\neg$ $idx$ \textbf{in} $executing\_child$}
        \State $child\_status \gets child(idx).Tick()$
    \Else
        \State $child\_status \gets child(idx).Status()$
    \EndIf
    \If {$child\_status ==$ \emph{FAILURE}} 
        \State \textproc{clear}($executed\_child$)
        \State \textbf{return} \emph{FAILURE}
    \ElsIf {$child\_status ==$ \emph{SUCCESS}} 
        \If {$\neg$ $idx$ \textbf{in} $executed\_child$}
            \State \textproc{ADD}($idx$ \textproc{in} $executed\_child$)
        \EndIf
        \If {$executed\_child.size() == child.size()$}
            \State \textbf{return} \emph{SUCCESS}
        \EndIf
    \EndIf
\EndFor
\State \textbf{return} \emph{RUNNING}
\EndProcedure
\Statex
\end{algorithmic}
\vspace{-3mm}
\end{algorithm}

\subsubsection{Agent Handler Node}
This node is defined as a custom condition node and it is in charge of selecting the agent, according to the allocation result. Specifically, the \textit{Agent Handler} reads the results and returns different status in case the task has to be communicated to a human worker or it has to be accomplished by a robot.
Thanks to the structure of the \textit{Planning and Allocation Handler} subtree, if the node returns \textit{FAILURE} the BT ticks the \textit{Human Communication} node that is in charge to communicate the allocated action to the human. While, if the node returns \textit{SUCCESS}, the BT ticks the \textit{Robot Action} nodes that make the robot executing the scheduled tasks. The pseudocode of the \textit{Agent Handler} node is summarized in Algorithm~\ref{algo:AgentHandler}.

\alglanguage{pseudocode}
\begin{algorithm}[!ht]
\small
\caption{Tick() function of the "AgentHandler" node.}
\label{algo:AgentHandler}
\begin{algorithmic}[1]
\Procedure{$\textproc{AgentHandler::tick}$}{$ $}
\State $getInput([W,\;A])$
\For{$[w,\;a]$ \textbf{in} $[W,\;A]$}
    \If {$w.type ==$ HUMAN}
        \State \textbf{return} \emph{FAILURE}
    \ElsIf {$w.type ==$ ROBOT}
        \State \textbf{return} \emph{SUCCESS}
    \EndIf
\EndFor
\EndProcedure
\Statex
\end{algorithmic}
\vspace{-3mm}
\end{algorithm}

\subsubsection{Robot Action Nodes}
The action nodes are in charge to trigger the activation of specific motions of the robots. The node communicates directly with the robot motion planner or with the controller, depending on the development of the node itself. In our case, we defined a finite set of actions, that represent motion primitives e.g. grasp, move, etc. The advantage to have primitive actions as nodes is that we do not need to create an action for each robot or different specifications of the primitive itself. A motion primitive has an interface, where it is possible to define all the specific information needed to be executed, position in space, force to be exerted, etc but then it is implemented differently in each robot. Another advantage consists in the fact that this action info can be used not only in the execution phase, but also by the role allocator node to compute the execution costs. When the agent starts the action execution, the node changes the worker availability and outputs it in the blackboard. Each agent has its own port to allow the execution of different actions in parallel avoiding more than one node changing the status of the same worker.  

\subsubsection{Human Communication Node}
This node is the corresponding node to the \textit{Robot Action} for the human workers.
Therefore, the \textit{Human Communication} node is in charge to communicate to each human agent the action is asked to execute by reading the allocation results.
The communication ways can differ in relation to the environment and/or the workers' equipment, e.g. displaying the information in a monitor, communicating it through wearable devices such as smartwatches or mixed reality smartglasses~\cite{lamon2019capability}.

\section{ROLE ALLOCATION}
\label{sec:role_allocation}
\subsection{Problem Statement}
The multi-agent task allocation (MATA), considered as a more general class of the well-known multi-robot task allocation (MRTA), is the problem of determining which agent, either human or robotic, is in charge of executing each single task that is needed to achieve the team's goal.

Before formalizing the role allocation problem, we need to specify which are the main components of the problem.
Following the symbolism introduced in~\cite{lamon2019capability}, we consider a mixed human-robot team of workers, or agents, $W=\{w_1,...,w_n\}$, $|W| = N$. The goal is to complete a general single job $A$ that could be further decomposed into the sequence and parallel of tasks, or actions, $A=\{a_1, .. ,a_m\}$, $|A| = M$. The set of $L$ actions that can be executed by each agent $w_i \in W$ are $A_i=\{a_{i1}, .. ,a_{il}\}$, $|A_i| = L$, where $A_i \subseteq A$. We want to obtain is the allocation of an agent $w_i$ to each of the actions he is able to execute $a_{ij}$, denoted $w_i \rightarrow a_{ij}$. The set of worker-action allocations is denoted $W \rightarrow A$.

\subsection{Mathematical Model}
In this work, the MATA problem, is formalized as a Mixed-Integer Linear Program. The most general scenario of MATA problems in human-robot collaboration, according to an adapted version of the taxonomy presented by \textit{Gerkey and Matari\'c}~\cite{gerkey2004formal}, is characterized by multi-task agents, i.e agents that can execute multiple tasks simultaneously, multi-agent tasks, i.e, tasks which requires multiple agents, and, finally, time-extended assignments, i.e. the allocation considers also future allocations. To the best of the authors' knowledge, a mathematical model that captures the features of such complex problem is still missing in literature~\cite{korsah2013comprehensive}.
However, in our framework, thanks to the decomposition of the task that BTs are able to achieve, the problem is extremely simplified. First, each agent, by definition, can perform only a task at once; second, each task require a single agent. Moreover, collaborative tasks, that require more than a single agent at the same time, are already decomposed by the BT into the parallel of tasks. By doing so, we have removed all the \textit{complex} and \textit{cross-schedule} dependencies, i.e. the effective cost of a agent for a task and the allocation constraints do not depend on the schedules of other agents.  
In practice, the BT \textit{Allocator} node deals only with the allocation of a sub-set of tasks, that are only the tasks that, according to the job schedule represented by the BT, should be allocated within the available agents, having as constraints, in the worst-case scenario, \textit{intra-schedule} dependencies, i.e. the agent cost for an action depends on the other actions the agent is performing.
Hence, the problem of allocating $L$ actions, where $L \le M$, to $N$ workers, can be formalized in the following way.

\vspace{1mm}
Minimize:
\begin{equation}\label{eq:role_allocation_cost}
   \sum_{w_i\in W}\sum_{a_j\in A}( c_{ij} + \chi_i)x_{ij} 
\end{equation}

Subject to:
\begin{equation}\label{eq:role_allocation_constr}
    \begin{aligned}
    x_{ij} & \in\{0,1\} & \forall w_i \in W, \forall a_j \in A \\
    x_{ij} & = 0 & \text{ if } a_j \notin A_i\\
    \sum_{w_i \in W} x_{i j} & \le 1 & \forall a_j \in A \\
    \sum_{a_j \in A} x_{i j} & \le 1 & \forall w_i \in W \\
    \sum_{w_i \in W} \sum_{a_j \in A} x_{i j} & = \min{(L, N)} \\
    \sum_{w_i \in W} \sum_{a_j \in A} t_{i j} x_{i j} & \leq T_{k} & \forall k \in K \\
    \end{aligned}
\end{equation}

\noindent where $c_{ij}$ represents the cost related to an agent $w_i$ in executing the task $a_j$, $\chi_i$ is the availability cost function, and the $x_{ij}$ represents the $M \times N$ binary optimisation variables of the problem, where $x_{ij} = 1$ means that the worker $i$ is assigned to action $j$ ($w_i \rightarrow a_{j}$).

\subsubsection{Constraints Design}
The problem constraints are exploited to ensure the feasibility of the problem: 
\begin{itemize}
    \item the first is representative of the binary nature of the variables;
    \item the second one ensures that the solver does not allocates a worker to an action that is not capable to execute. Another, more common, solution, consists in penalizing the allocation by assigning a large cost for the task;
    \item the third and fourth constraints ensures that to each agent only one task is allocated, and the same task is not allocated to two different agents.
    \item the fifth constraint ensures that there are exactly a number of allocations equal to the number of agents $N$, in case $N > L$, where $L$, are the tasks to be allocated, and $L$ otherwise;
    \item the budget constraint ensures limits the number of tasks assigned to each agent, where $t_{ij}$ is the budget that $w_i$ would spend for $a_i$, and $T_k$ is the budget limit for the $K$ joint agent-task constraints.
\end{itemize}

\subsubsection{Costs Design}
\label{subsec:costs_design}
In this work, the optimization costs are split into two main components: the agent-actions related costs $c_{ij}$ and the agent-related availability cost $\chi_i$. The first should describe how good is a worker in performing each tasks, considering the kino-dynamics features of the agents, the tasks duration, the human ergonomics etc. The role and the design of the costs for such problems has already been described in our previous work~\cite{lamon2019capability} and, hence, it will not be repeated here.

The availability value has been considered into the cost function, instead of into the problem constraints, since we cannot ensure that at least an agent is always available. In case no agents, or multiple ones, are available, the allocator node tries to assign the task to the agent whose cost is smaller.
We consider an agent available if it is present in the workcell and if it is not occupied by any other task. One simple definition of the availability activation is the following:
\begin{equation} \label{eq:binary_availability}
    \chi_i = \begin{cases}
    0, \quad &\text{if } w_i \text{ is } \textit{available};  \\
    \alpha_i, \quad &\textit{otherwise.}
    \end{cases}
\end{equation}
$\alpha_i$ is the availability cost of $w_i$. To ensure that the availability weights more than the other costs, we set $\alpha_i > \max{\{c_{ij}\}}$. With this method, we are able to ensure that the allocation node favors an available agent, minimizing in this way the agents waiting times. In this case, the agent suitability to the task is not considered. On the other hand, in some situations, instead allocating a task only among the available agents, that might all be strongly unsuitable for the task, it might be convenient to make the system to wait for the most suitable agent. For this reason, we modified the binary nature of the availability to account for the remaining execution time:
\begin{equation} \label{eq:availability}
    \chi_i = \begin{cases}
    0, \quad &\text{if } w_i \text{ is } \textit{available};  \\
    \alpha_i \frac{T_{a_{ij}} - t_{a_{ij}}}{T_{a_{ij}}} , \quad &\textit{otherwise.}
    \end{cases}
\end{equation}
where $T_{a_{ij}}$ is the nominal duration of $a_j$ performed by $w_i$, $t_{a_{ij}}$ is the time spent by $w_i$ for $a_j$ from the beginning of $a_j$, where $t_{a_{ij}} \le T_{a_{ij}}$. In this way, $t_{a_{ij}} = 1$ if the task has just started, and, as the agent executes the task, it goes to 0.  
To ensure that the two costs are comparable, $\alpha_i = \max{\{c_{ij}\}}$.

\section{EXPERIMENTS}
The performances of the proposed approach are evaluated through simulation experiments. In order to analyze the performances of the method, different experiments are conducted varying the job characteristics, such as the number of actions in sequence, in parallel and the number of workers.
The simulation experiments were run on a laptop with an Intel Core i7-8565U 1.8 GHz $\times$ 8-cores CPU and 8 GB RAM. The architecture has been developed in C++, on Ubuntu 18.04 and ROS Melodic, exploiting the BehaviorTree.CPP (\url{https://github.com/BehaviorTree/BehaviorTree.CPP}) library to define the BT nodes and the Osi (\url{https://github.com/coin-or/Osi}) library with GLPK (GNU Linear Programming Kit) solver to formalize the MATA problem.

\begin{figure}
    \centering
    \includegraphics[trim=11cm 1cm 13cm 3cm,clip,width=\linewidth]{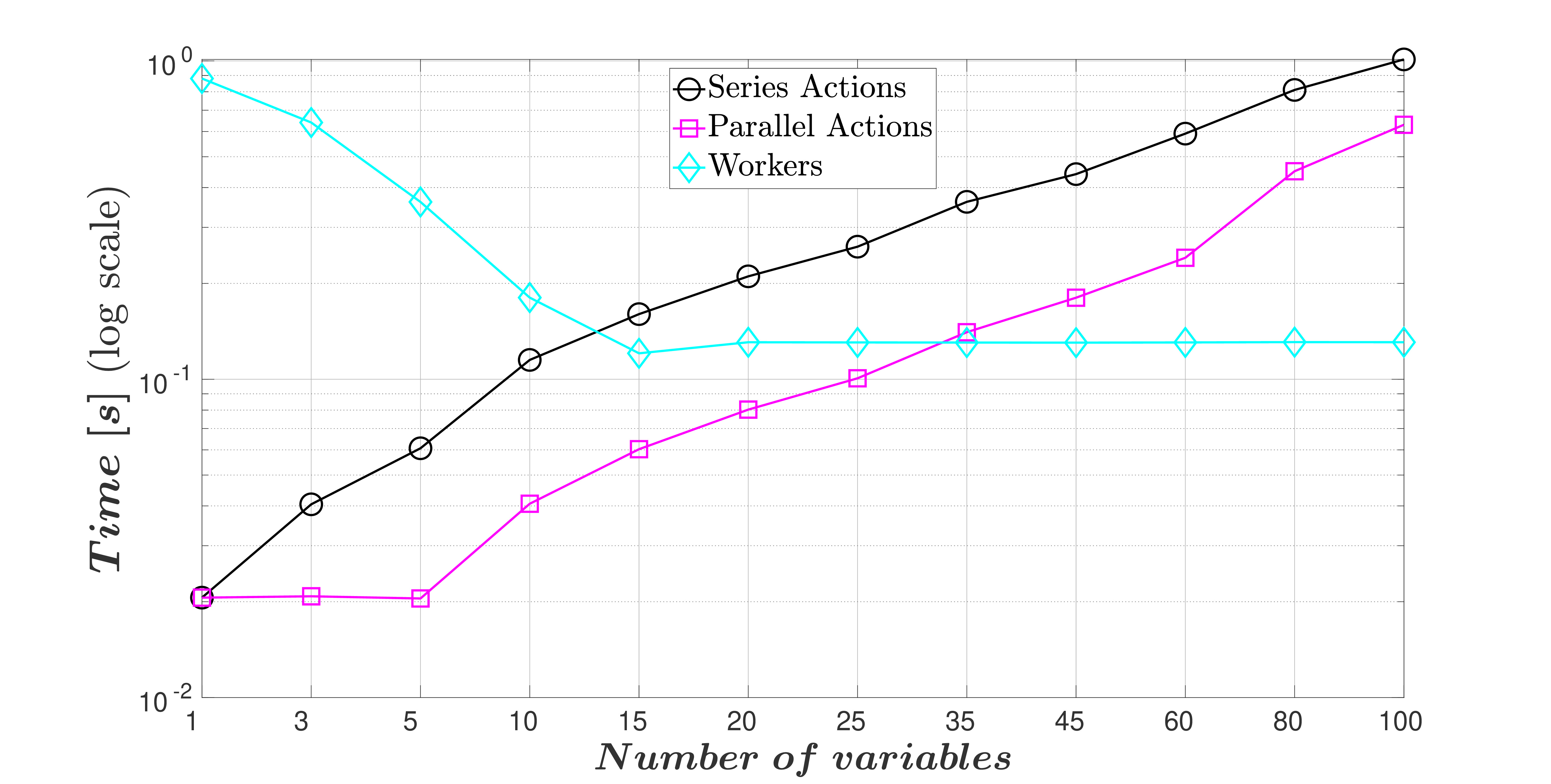}
    \caption{Computational time (in log scale) of series (black) and parallel (pink) actions and workers (cyan), with different number of variables.}
    \label{fig:computational_time}
    \vspace{-6mm}
\end{figure}

First, the computational complexity of the whole proposed method is evaluated. The computation time has been counted from the initialization of the BT until the action execution, excluded the execution time of the action, as the mean of 10 repetitions of the same BT. 
To generate different situations, the number of tasks in sequence, in parallel and the number of agents is progressively increased. The tasks in parallel are all children of the same allocation node, hence increasing parallel tasks means increasing the size of the MILP. On the other hand, the tasks in sequence are not in parallel with any task, and, hence, increasing the number of tasks in sequence means increasing the number of MILPs to solve. While the number of tasks increases, the number of workers is fixed to 4. Both actions in parallel and in series are increased from 1 to 100.   
The results in \autoref{fig:computational_time} show that the trend of the computation time increasing both series and parallel actions can be approximated with a linear function. It is interesting to notice that, even if the trends are similar, in this specific scenario, it is faster to solve a $N$-sized MILP than $N$ MILPs. This, however, does not mean that solving the offline centralized problem is faster than the decomposed sub-problems since the centralized one should include all the task temporal constraints.  
A different simulation is conducted to estimate the evolution of the computation time increasing the number of workers keeping the number of actions fixed. In this case, the BT is constructed to have 11 \textit{Role Allocator} nodes in sequence, each one with: 12, 8, 10, 12, 8, 10, 12, 8, 10, 5, 5 number of parallel actions, respectively. Again, the number of workers is increased starting from 1 until 100, and, as can be noticed in \autoref{fig:computational_time} in cyan, increasing the number of agents the computation time decreases. This is because the number of actions executed in parallel increases with the workers until the maximum degree of parallelism is reached. Specifically, this happens when the agents are 15 and, hence, are more than the maximum number of actions in parallel, i.e. 12. From that value on, the computation time is approximately constant.

\begin{figure}
    \centering
    \includegraphics[trim=15cm 3cm 3.5cm 5cm,clip,width=\linewidth]{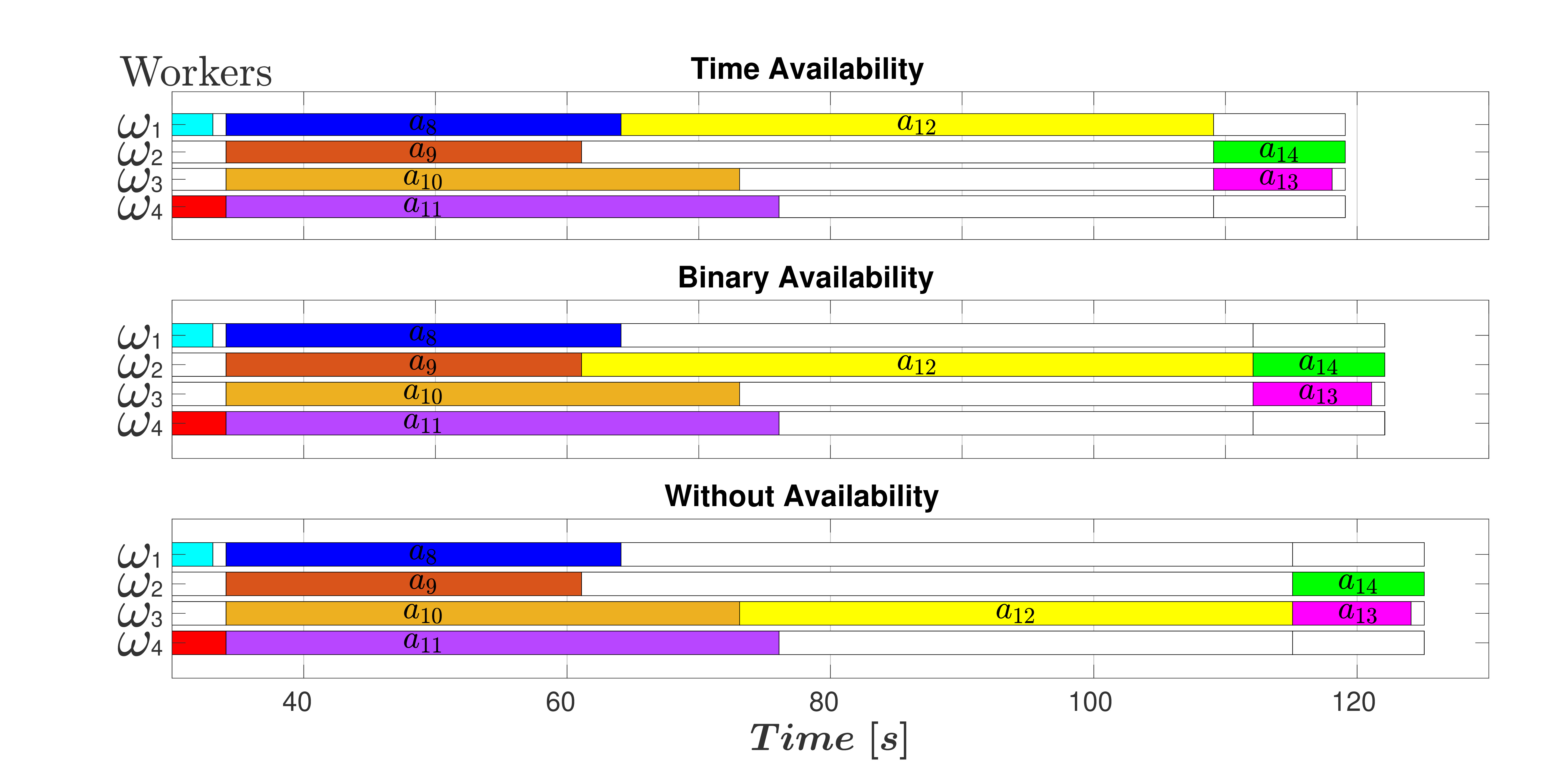}
    \caption{Detail of the Gantt charts related to the allocation of actions $a_8-a_{12}$: in the top plot the availability cost $\chi_i$ is computed as in \autoref{eq:availability}, in the middle one as in \autoref{eq:binary_availability} and in the bottom one $\chi_i = 0$.}
    \label{fig:gantt_chart}
    \vspace{-4mm}
\end{figure}

Then, the allocation approach is validated with a single fixed job. The goal is to run the allocation method for the same job with different costs, to validate the different design of the availability cost $\chi_i$ explained in \autoref{subsec:costs_design}.
The job is made of a total of 14 actions, combined in a sequence of 4 different sets of actions that can be executed in parallel. For this reason, 4 \textit{Role Allocator} nodes are present in the BT. Each node has 3, 4, 5 and 2 actions as children, respectively. The number of workers is fixed to 4.
The agent-actions related costs $c_{ij}$, for simplicity, are defined as the time ($t_{w_i}$) required by the worker $w_i$ to execute the action $a_j$.
First, we compute the allocation with availability cost defined in \autoref{eq:availability}. The results in \autoref{table:allocation_results} show that, in general, the algorithm picks always the agent that minimizes the execution time. These results are then compared with the same allocation method computed with the binary availability  \autoref{eq:binary_availability} and without any availability cost ($\chi_i = 0$). The results differ for the allocation of action $a_{12}$ (see \autoref{fig:gantt_chart}).
Without considering the availability of the agents, the action $a_{12}$ is assigned to the worker $w_3$ minimizing only the agent-actions related costs $c_{ij}$. This, however, is not optimal since $w_3$ is occupied by another task. In the case of the binary availability, the action $a_{12}$ is assigned to the worker $w_2$, since he's the first available after finishing $a_{10}$. But, even if the $w_2$ starts before, the execution time required by him to achieve $a_12$ is higher than the expected waiting time for the execution of $a_8$ by $w_1$ plus the following $a_12$, that is the solution proposed with availability cost in \autoref{eq:availability}. Consequently, this solution minimizes also the overall duration of that set of tasks, and also the waiting time for the other agents for the next task.   
It can be noticed that actions $a_8$, $a_9$, $a_{10}$ and $a_{11}$ starts at the same time because are defined as parallel actions but in series with the previous ones, hence, these can start only when all the previous ones are completed.

\section{CONCLUSIONS}
In this work, we proposed a novel integrated method to allocate and plan tasks for mixed human-robot teams. The method extends the standard formulation of BT with custom reusable nodes that enable to dynamically generate and solve different MILP sub-problems. The results showed the crucial role of the cost definition in the allocation behavior. For these reasons, further metrics should be evaluated with the method, to reduce the agent workload by optimizing the human ergonomics.  
Furthermore, future studies will compare the method with other state-of-the-art approaches, focusing not only on the computational complexity but also on the intuitiveness and user-friendliness assessment of the interface to generate the jobs. Finally, the effectiveness of the allocation method should be evaluated with real experiments in multi-human and multi-robot teams. 

\begin{table}[!t]
\caption{Allocation results of the actions of the simulated job.}
\vspace{-4 mm}
\label{table:allocation_results}
\begin{center}
\begin{tabular}{|c||c|c|c|c|c|}
\hline
\textbf{Action} & $\boldsymbol{t_{w_1}} (s) $ & $\boldsymbol{t_{w_2}} (s) $ & $\boldsymbol{t_{w_3}} (s)$ & $\boldsymbol{t_{w_4}} (s)$ & \textbf{Worker allocated} \\[0.5mm] 
\hline
$a_1$ & 20 & 13	& 15 & 15 & \cellcolor{blue!50} $w_2$\\
$a_2$ & 17 & 20	& 22 & 16 & \cellcolor{green!50} $w_4$\\
$a_3$ & 10 & 12	& 17 & 11 & \cellcolor{orange!50} $w_1$\\
$a_4$ & 13 & 15	& 9 & 21 & \cellcolor{purple!50} $w_3$\\
$a_5$ & 22 & 18	& 24 & 18 & \cellcolor{green!50} $w_4$\\
$a_6$ & 11 & 9 & 15 & 15 & \cellcolor{blue!50} $w_2$\\
$a_7$ & 17 & 23	& 18 & 16 & \cellcolor{orange!50} $w_1$\\
$a_8$ & 30 & 54	& 48 & 57 & \cellcolor{orange!50} $w_1$\\
$a_9$ & 66 & 27	& 60 & 39 & \cellcolor{blue!50} $w_2$\\
$a_{10}$ & 60 & 66 & 39 & 75 & \cellcolor{purple!50} $w_3$\\
$a_{11}$ & 63 & 48 & 57 & 42 & \cellcolor{green!50} $w_4$\\
$a_{12}$ & 45 & 51 & 42 & 54 & \cellcolor{orange!50} $w_1$\\
$a_{13}$ & 14 & 17 & 9 & 16 & \cellcolor{purple!50} $w_3$\\
$a_{14}$ & 21 & 10 & 15 & 18 & \cellcolor{blue!50} $w_2$\\
\hline
\end{tabular}
\end{center}
\vspace{-7mm}
\end{table}

\bibliographystyle{IEEEtran}
\bibliography{biblio.bib}

\end{document}